# A Discussion on Influence of Newspaper Headlines on Social Media

Aneek Barman Roy
School of Computer Science and Statistics
Trinity College Dublin, University of Dublin
Dublin, Ireland
barmanra@tcd.ie

Baolei Chen
School of Computer Science and Statistics
Trinity College Dublin, University of Dublin
Dublin, Ireland
bachen@tcd.ie

Siddharth Tiwari
School of Computer Science and Statistics
Trinity College Dublin, University of Dublin
Dublin, Ireland
stiwari@tcd.ie

Zihan Huang
School of Computer Science and Statistics
Trinity College Dublin, University of Dublin
Dublin, Ireland
huangzi@tcd.ie

April 12, 2019

*Abstract*

*Newspaper headlines contribute severely and have an influence on the social media. This work studies the durability of impact of verbs and adjectives on headlines and determine the factors which are responsible for its nature of influence on the social media. Each headline has been categorized into positive, negative or neutral based on its sentiment score. Initial results show that intensity of a sentiment nature is positively correlated with the social media impression. Additionally, verbs and adjectives show a relation with the sentiment scores.*

## 1 Introduction

Headlines are based on facts and opinions where parts of speech play an important role in influencing a reader's mind. Most headlines generate varied levels of curiosity depending upon the content [1]. Any particular news has the ability to generate a positive or negative impression. Irrespective of the nature of the content, strong reactions lead to more dialogues and subsequently more user engagement. Presently, more efforts are put to increase engagement on social media [2]. Interestingly, many companies are trying to build their brands through value creation and tailored news [3]. However, a gloomy article can create a negative impression amongst the readers. Flower mentioned in his book [4] how the use of verbs is an essential characteristic of print headlines. However, writers often use strong words as an objectivity norm [5] to make headlines more appealing to the readers.

This has led us to make two hypotheses:

H1: Verbs have greater impact than Adjectives in a print headline.

H2: Print headlines of companies has an influence on social media

For H1, we used parts of speech tagging technique to determine the position of verbs and adjectives with respect to other parts of speech viz pronouns, adverbs, nouns etc, in a headline. A sentiment analysis was directed on headlines to generate opinions and consequently the impact of verbs and adjectives were measured with respect to the sentiment score. Additionally, we try to find, reason and eventually discuss on how largely can a newspaper headline have an influence on social media. Henceforth, to validate H2, we have proceeded for the approach through companies which were mentioned in the business section of print media throughout a month. The number of mentions throughout various social media platforms for such companies were tracked and correlated with respective sentiment score.

The corpus for the print media headlines was compiled from the publications of New York Times from February 20, 2019 to March 19, 2019. Since our target actors include multinational companies, we have chosen the business section of the publication.

A literature review of related journals and previous works have been discussed in Section 2. The methodology of proposed research and related results and discussions is present in Section 3 and Section 4 respectively. Finally, we conclude our paper with outcomes and future work.

## 2 Related Work

Paula Chesley [6] et. al classified blog sentiments automatically using verbs and adjectives where each post was considered to be objective, positive or negative. After the information was retrieved using Semantex, the classifier Support Vector Machine (SVM) was assigned to each of the post. While approving and asserting verb classes contributed severely to the accuracy of positive classification, the work additionally implied that the verb classes can refine results on sentiment classification.

Karamibekr and Ghorbani [7] proposed a new approach based on the verb being used as an essential opinion term. Their work, based in social domains, extracted opinion structures and then recognized their orientations regarding the social issue. However, the authors did not correlate the sentiment of opinion structures with the sentiments of other structures. Additionally, use of synonyms and antonyms could have been a good approach to solve the issue and improve the algorithm.

SVM [8] provide the best performances during classification of the sentiment of a document. Interestingly, Chesley, Karamibekr and Ghorbani used SVM to run their semantic analysis on verbs whereas Hatzivassiloglou and McKeown [9] went with a model of log-linear regression to determine the semantic orientation of adjectives that are conjoined. However, the classified subsets of adjectives having different orientation using a clustering algorithm and concluded a better accuracy score. The research work did not extend the analyses to adjectives or verbs.

Benamara [10] et. al. tried to use adjectives and adverbs and proposed a linguistic approach to sentiment analysis where -1 and +1 represented negative and positive opinion respectively. The authors implemented a linguistic classification of adverbs of degree (AoD) by defining general axioms. In contrast, our paper focuses on the necessity of adjectives and verbs. In their esteemed work, Ong Hui [11] et. al. suggested that news and word classes play vital roles which affect the process of news classification. The study took into account on how the word positions can exert influence on a reader's sentiment. The authors used WEKA for the classification process and subsequently implemented supervised learning.

## 3 METHODOLOGY

A headline can make a positive or negative impact on the social media. To answer the research topic i.e. A Discussion on Influence of Newspaper Headlines on Social Media, a relationship between the sentiment score of a headline was established with impact of the company on social media. An outline is given in Figure 1.

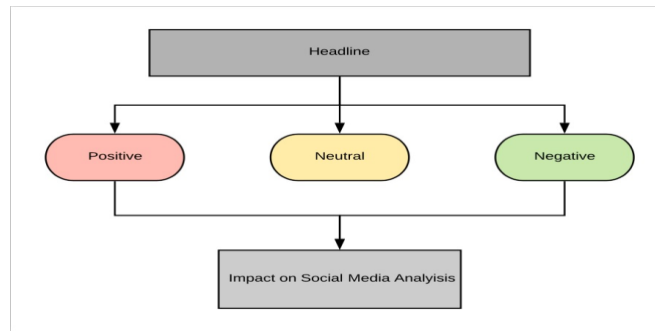

Figure 1: Generic Model of How Headline Can Influence Social Media Analysis

To answer H1, the frequency score of verbs and adjective was measured and its adverse effect on the sentiment score of a headline.

A novel system has been generated to calculate the total sentiment score of each company in a month and relate with its social media mentions. The process is carried out in three levels namely – document level, sentence level and opinion level.

Document Level - Each headline is loaded into the document level where sentences are checked and separated in case of the presence stop words in a headline.

- Document Level – Each headline is loaded into the document level where sentences are checked and separated in case of the presence stop words in the middle of a headline.
- Sentence Level – Each sentence is then processed and a sentiment score ($S_i$) is generated

$$S_i = \text{Sentiment Score} \quad \text{where } i = 0, 1 \ \{0 \sim Negative, 1 \sim Positive\}$$

- Opinion Level – Based on i, the sentiment is categorized into positive or negative. If i=0, it is categorized as neutral in nature

Once the $S_i$ is obtained, the total sentiment for each company is calculated by the formula:

$$\text{Total Sentiment} = \sum_{j=1}^{n} positiveSentiment + \sum_{j=1}^{n} negativeSentiment$$

where $S_0$ = negativeSentiment and $S_1$ = positiveSentiment

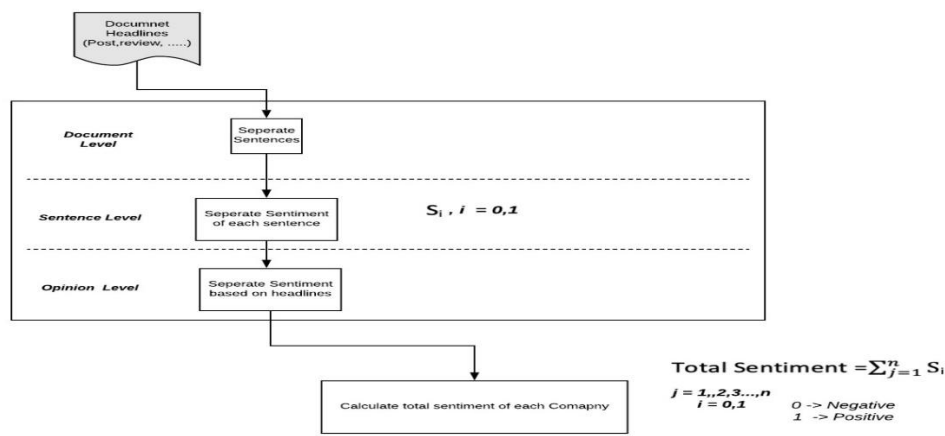

Figure 2: A System to Calculate Company Sentiment Score

The date of publishing any news article is important as the impact factor on that particular day is more than the following days. The impact of various articles tends to die down on the next day. However, if user engagement is more in the subsequent days, it would suggest a positive impact. The number of social media mentions is a good measure to determine the user engagement of any company [12]. A trend of number of mentions of any company over a period of five days could be negative or positive depending on its tendency to decrease or increase.

To lay a validation to H2, a positive impact will hold true if $d_1 < d_{average}$, where $d_1$ is the number of mentions of a company on the first day and $d_{average}$ is the average of the total number of mentions of a company for a period of five days. A positive impact will happen if *impact=1* whereas it will be negative in nature if *impact=0*. The process is outline in *Figure 4*.

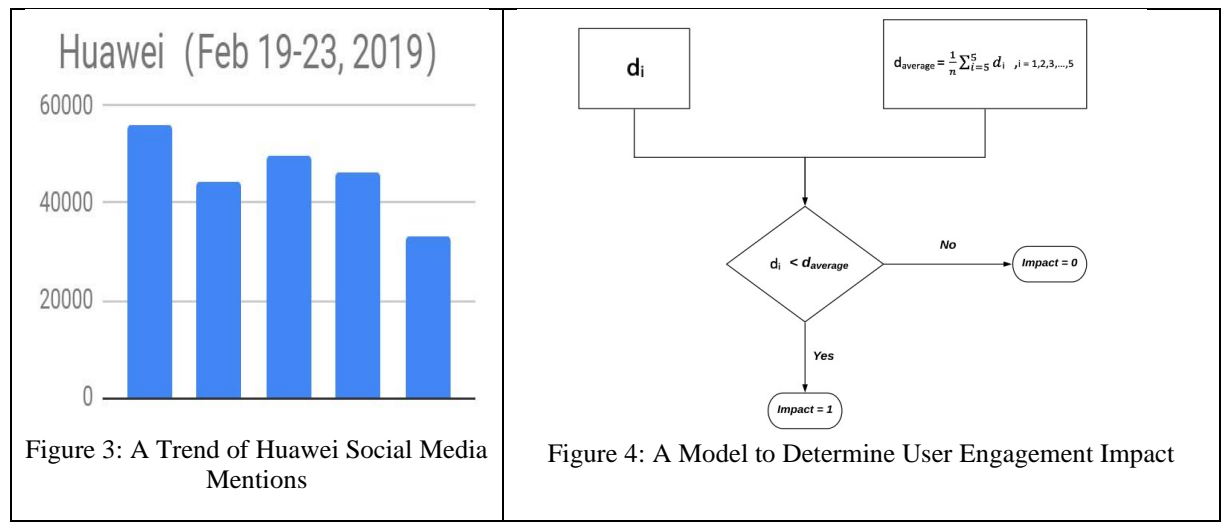

Figure 3: A Trend of Huawei Social Media Mentions

Figure 4: A Model to Determine User Engagement Impact

A major portion of our research work is centred around text and sentiment analysis. Besides this, we have used a novel approach to correlate the analysis from headlines with that of the mentions on social media. The collection, pre-processing and analysis has been discussed below.

### 3.1 Data Collection and Extraction

Lexis Nexis [13] is an electronic recourse database available as an online recourse through the college library. The database houses articles and snippets from dozens of popular newspapers like the Washington Post, Irish Times, New York Times etc. For this project, we gathered headlines from the business section of New York Times.

Initially, we scraped data of New York Times from February 20, 2019 to March 19, 2019 and collected a total of 3370 headlines. However, there were a lot of headlines which did not belong to the business section. Thus, we had to pre-process the data using a statistical analysis tool – R.

Based on the mention of a company per day in a headline, we gathered social media statistics of that particular company from talkwalker.com - an analysis and monitoring platform for social media [14]. It was collected on a regular basis where we put focus on its number of mentions across social media platforms for a period of five days from the date of publishing of each headline.

### 3.2 Data Pre-processing

The obtained dataset from the corpus was initially cleaned using various techniques viz. removal of duplicate, noisy, unnecessary data. Subsequently, we selected only the data where business category was present. On obtaining 237 of such records, we found out that there were 199 cases where the name of a company was present in a headline.

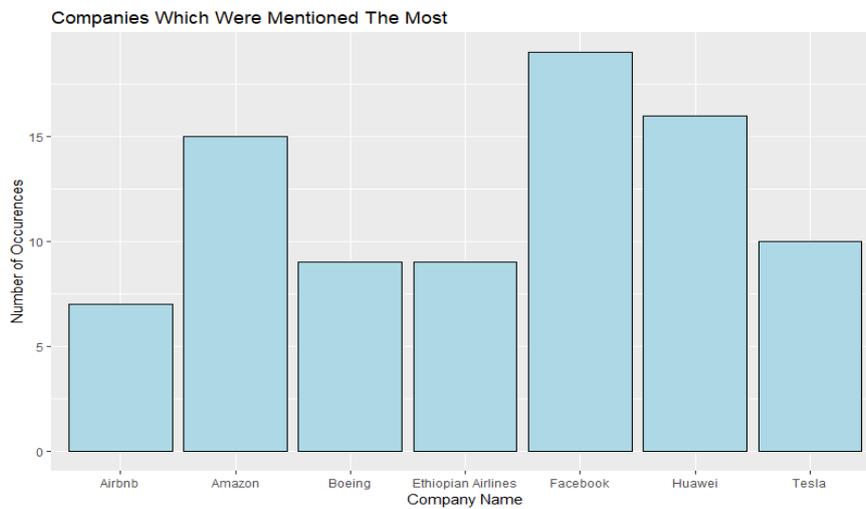

Figure 4: Top 7 Companies with Most Headlines

## 3.3 Text Analysis

The core concepts of text analytics were used to run several analyses. A trained linguistic model of udpipe [15] was run on the data set which took each headline at a time, measured and returned several parameters viz. tokenization, POS tags etc. Since our headlines were from an English daily, we downloaded and ran the English based model of the package.

### 3.3.1 Textual Features

The model returned statistics of different parts of speech along with other functionalities. It has been found that punctuation has the highest frequency percentage score. This infer that punctuation occur very frequently followed by ad position (ADP) and noun. Interestingly, the frequency percentage score of verbs is more than double of the score of adjectives.

| KEY | FREQ_PCT |
|---|---|
| PUNCT | 9.41 |
| ADP | 8.77 |
| NOUN | 8.58 |
| VERB | 5.54 |
| DET | 4.07 |
| PART | 3.67 |
| PRON | 3.04 |
| AUX | 2.69 |
| ADJ | 2.15 |
| CCONJ | 1.96 |
| NUM | 1.71 |
| ADV | 1.37 |
| SYM | 0.68 |
| SCONJ | 0.53 |
| INTJ | 0.19 |
| X | 0.04 |

Table 1: Frequency Scores of POS

### 3.3.2 Feature Selection:

Our hypothesis(H1) enables to work on findings which involve verbs and adjectives. With both adjectives and verbs having frequency scores greater than 2.0, it becomes imperative to select them to determine their relationship with the sentiment of a headline. Based on observations, it can be said if the impact of verbs is more than adjectives or vice versa.

## 4 RESULTS AND DISCUSSION

This section comprises of findings which were realized during the research. Initially, the number of mentions of the companies across several social media platforms were analysed followed by the analysis of headlines. The sentiment of each headline was then analysed and specific relationships of headline sentiments were established between verbs and adjectives and the company social media mentions. Finally, the model was evaluated which helped in addressing the conclusion of the whole discussion.

### 4.1 Analysis of Social Media Mentions

The dataset compiled from the data of talkwalker.com [14] was cleaned and duplicate or missing values were eradicated. This has resulted in providing promising results. It was discovered that companies having higher number of mentions on the publishing date were mostly social media platforms and this trend is almost similar across the initial five days from the date of publishing. This could be due to the fact that social media platforms engage users on a daily basis. Henceforth, such an analysis will not help to remove the existing bias and answer H2.

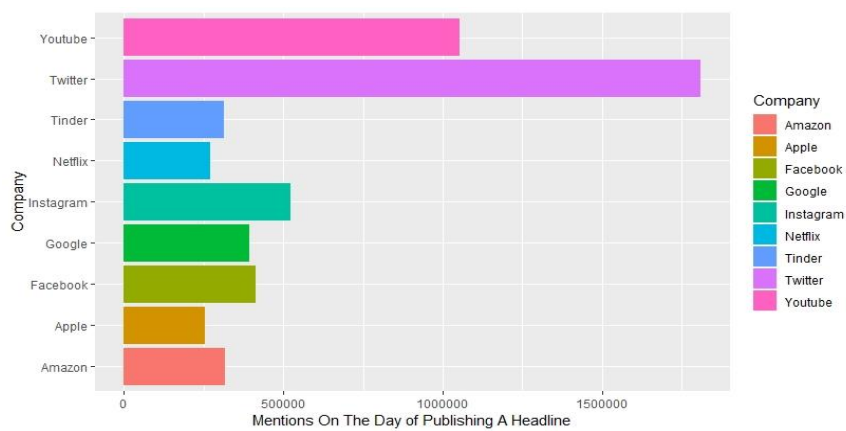

Figure 5: Companies Which Had Most Social Media Mentions On Publishing Date

It was due to this reason; the impact factored approach was designed and has been explained in the methodology section. The user engagement for a company after the publishing data will be positive if *impact = 1* as discussed in Figure 4. Additionally, this would imply that a company related headline having a positive impact will engage more users over the following days as the trend of the graph would tend to increase.

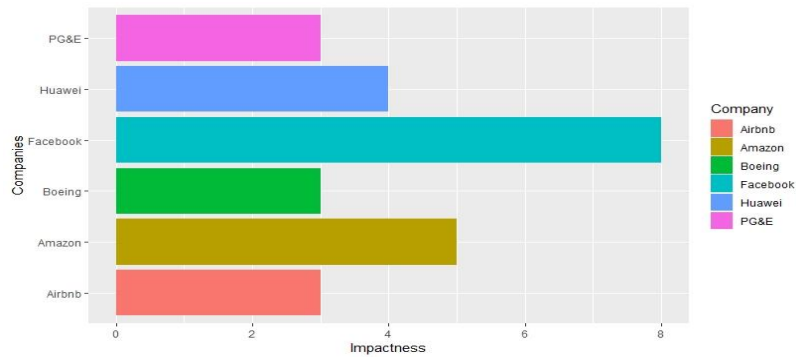

Figure 6: Companies with Most Impact

Figure 6 include the companies which had a higher impact score in terms of user engagement. The results can be correlated with the nature of headlines sentiments to answer our H2.

### 4.2 Analysis of Headlines

As discussed in the methodology, the dataset which has been pre-processed contain headlines of each company in a month. Initially, several analyses were implemented on all the headlines to determine the most occurring words.

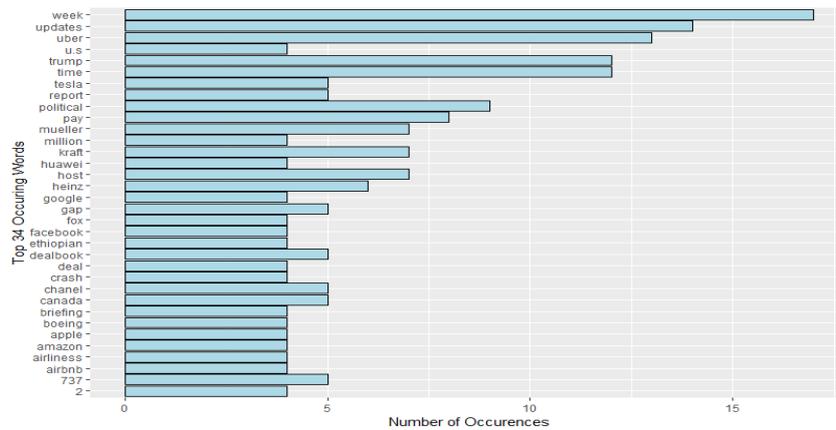

Figure 7: Most Occurring Words

Using Parts of Speech (POS) tagging techniques, the occurrence of verbs and adjectives along with their cooccurrence were derived from the dataset. The results have indicated that verbs and adjectives have a strong cooccurrence between them. The frequency score of verbs was found to be higher than that of adjectives.

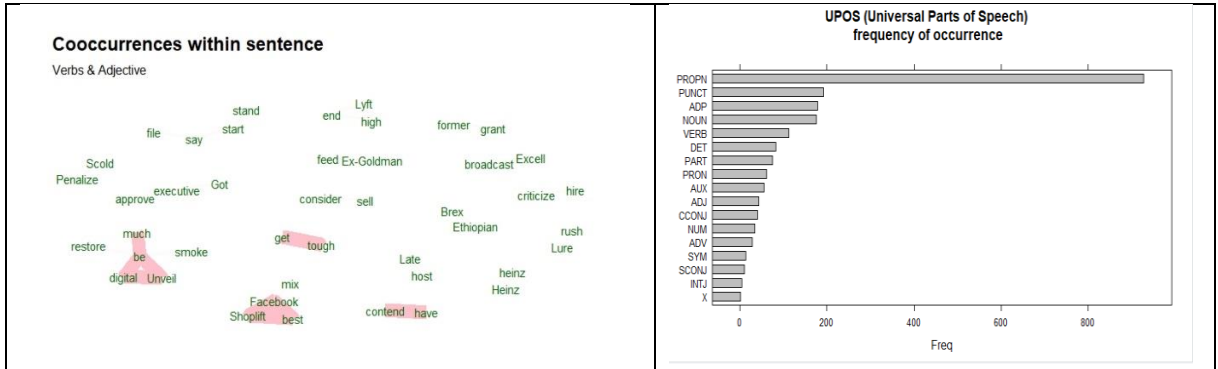

Figure 7. Occurrence and Cooccurrence of Verbs and Adjectives

The NRC sentiment test on each headline showed words which contributed to different emotions like anger, anticipation, disgust, fear, joy, negative, positive, sadness, surprise and trust. The presence of a larger number of sentiment indicators will give arise to a lot of complexities in terms of correlations and comparisons. Thus, there was a need to have a sentiment analysis where the headlines can be classified as negative, neutral or positive. As per methodology, if *sentimentScore <0,* the headline is negative in nature while it is positive if *sentimentScore >0*. The sentiment score on each headline is obtaining by running a model of udpipe [15] in R. A neutral headline does not have any influence as its sentiment score is zero.

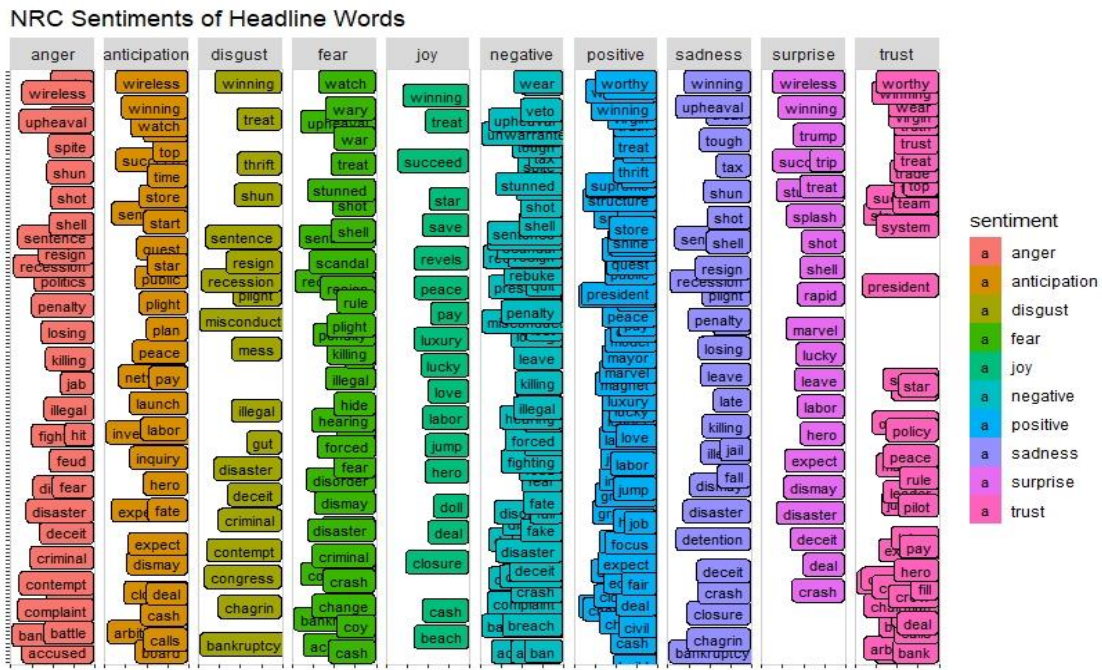

Figure 8: NRC Sentiment Distribution of Headline Words

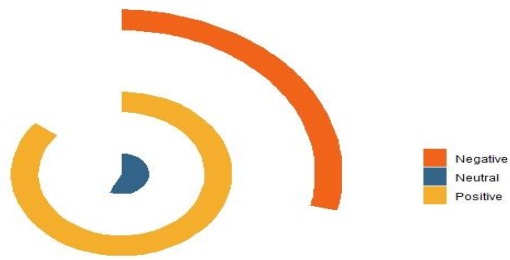

Figure 9: Sentiment Distribution of All Headlines

The most negative headline printed was pointed towards Times' -

`"Time's Up Chief Quit Over Sexual Misconduct Accusations Against Her Son"`

Total sentiment score for each company was derived through computation to determine the impact of such headlines on social media. The sentiment scores for companies with most headline occurrences is given below.

| Company | Total Sentiment Score |
|:---:|:---:|
| Boeing | 2.30 |
| Amazon | 2.10 |
| Facebook | -0.40 |
| Tesla | -1.00 |
| Airbnb | -3.20 |
| Huawei | -4.95 |
| Ethiopian Airlines | -5.35 |

Table 2. Sentiment Scores of Companies with Most Headline Occurrences

Interestingly, lesser number of negative headlines were published for Ethiopian Airlines while a larger number of positive headlines outnumbered the negative headlines for Facebook. This indicates that the intensity of a sentiment score is an important aspect for H2.

The most negative news for Ethiopian Airlines – "`Ethiopian Airlines Ethiopian Airlines Crash Updates: Pilots in U.S. Had Raised Concerns About Boeing 737 Max 8`" with a score of -1.35.

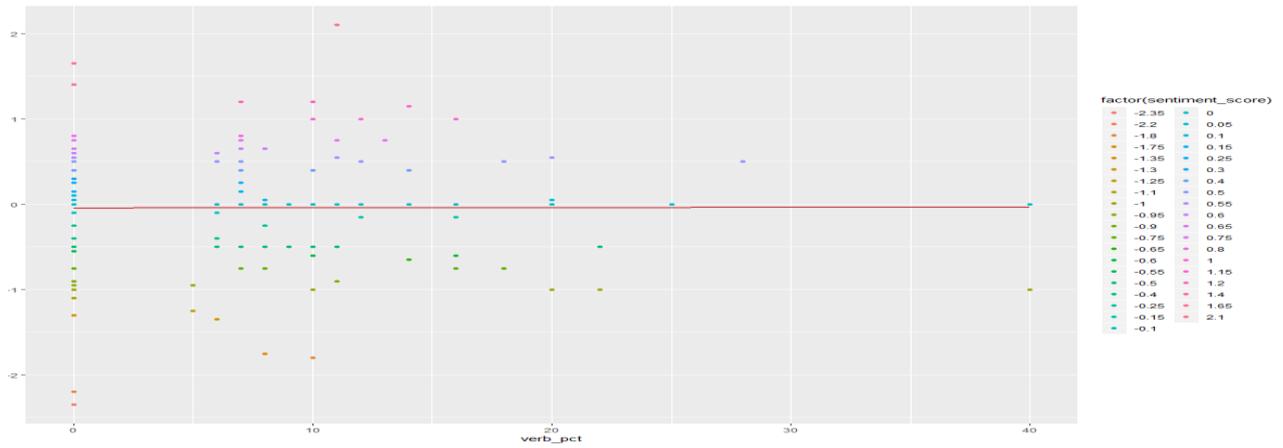
Figure 10. Impact of Verbs on Sentiment Score

From Figure 10 and Figure 11, it can be said that verbs have more density than adjectives with respect to the sentiment scores. Interestingly in the both the cases, the positive impressions were more than the negative impressions. Secondly, it can be inferred that verbs are more positively related with frequency score than adjectives. Thus, it can be said that verbs have more impact than adjectives. Figure 11 is presented below.

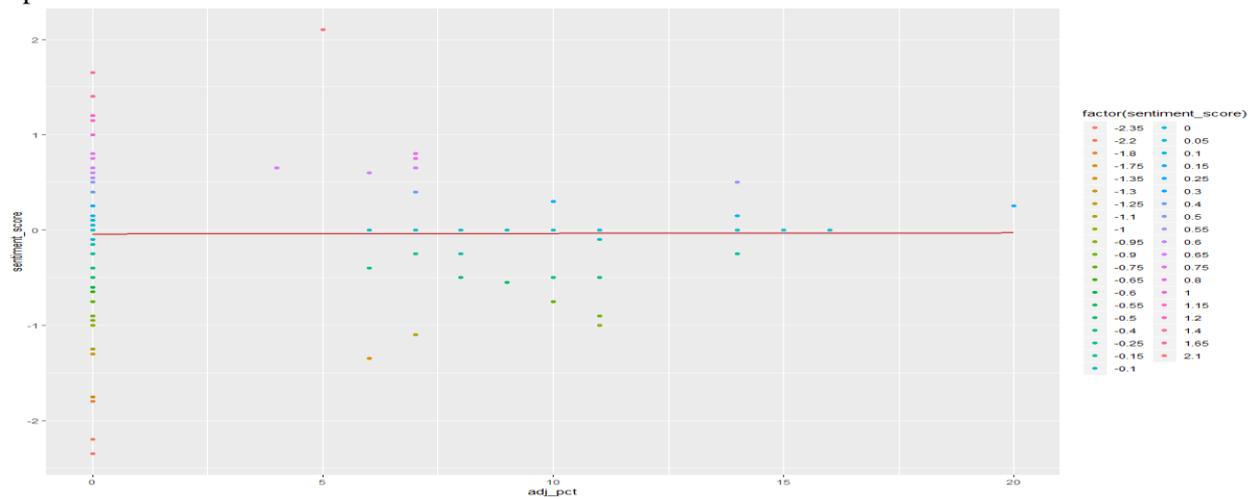
Figure 11. Impact of Adjectives on Sentiment Score

### 4.3 Evaluation of Model

The udpipe package in R comes with a function to check the accuracy. Since the research work involved usage of its linguistic model on POS tags, we ran the tagger performance evaluation on POS in two different settings: using raw text and using tokenization.

|             | Using Raw Text | Using Tokenization |
|-------------|----------------|--------------------|
| Iteration 1 | 37.04          | 48.15              |
| Iteration 2 | 81.48          | 77.78              |

Table 3: Accuracy Score of Model using Raw Text and Tokenization

### 4.4 Conclusion

For H1, the hypothesis holds true as the verbs have higher positive correlation with respect to frequency scores and positive impressions than adjectives. This proves that verbs have more impact than adjectives.

For H2, we focused on companies with higher number of print headline frequencies and companies with higher impact rate on social media platforms. Interestingly, most companies with high impact have higher number of headlines printed throughout the month. Additionally, companies with extreme sentiment scores have higher impact scores which means that extreme negative or positive reactions from headlines can lead to higher user engagement across the social media platforms. Thus, all these factors validate the other hypothesis(H2).

## 5 OUTCOME AND LIMITATIONS

It has been concluded that both hypotheses hold true. During the validation of H2, essential insights were found with respect to the impact of a headline on social media. Through the Results and Discussions section for the research topic, it has been observed that the impact score of a company on social media is positively related with the frequency of mentions in a printed daily. Additionally, it has been found that the intensity of a headline sentiment can affect the overall reaction of a company i.e. a higher positive headline can outweigh the impact of a lesser negative headline. Lastly, any headline will generate an impact if user engagement is more irrespective of the nature of headline.

Furthermore, the accuracy of udpipe model was used to evaluate the performance. Moreover, a correlation matrix could have been used to determine the co-occurrence of verbs and adjectives. A better approach would have been to evaluate in terms of precision, recall, accuracy and F1 score. The research work did not consider taking headlines from articles of different publishing house with varied reputation. As an able approach in future, it can be argued that how quickly different textual features in a headline can influence a positive impact on social media platforms.

## ACKNOWLEDGEMENT

This research work has been conducted under guidance and mentorship of Dr Carl Vogel. This essay is written as part of the 2018/19 Text Analytics Module (CS7IS4) at Trinity College, University of Dublin, Dublin, Ireland.[19]